\documentclass[a4paper]{article}
\usepackage[utf8]{inputenc}
\usepackage[T2A,T1]{fontenc
}\usepackage{multicol}

\usepackage{cite}
\usepackage{amsmath,amssymb,amsfonts}
\usepackage{algorithmic}
\usepackage{graphicx}
\usepackage{textcomp}
\usepackage{xcolor}
\usepackage{dot2texi}
\usepackage{tikz}
\usepackage{titlesec}
\usepackage{geometry}
\usepackage{paralist}
\usepackage{float}
\usepackage{hyperref}
\usepackage{libertine}
\usepackage{booktabs}

\newcommand{\ru}[1]{{\fontencoding{T2A}\selectfont
#1%
}}

\begin{document}

\title{Human-Annotated NER Dataset for the Kyrgyz Language
% \thanks{The work of Anton Alekseev was supported by the Russian Science Foundation grant \# 23-11-00358.}
}

\author{%
    Timur Turatali \\
    \textit{The Cramer Project}\\
    Bishkek, Kyrgyzstan \\
    timur.turat@gmail.com
    \and
    Anton Alekseev \\
    \textit{PDMI RAS, SPbU, KFU}\\
    St. Petersburg/Kazan, Russia \\
    \textit{KSTU n.a. I. Razzakov}\\
    Bishkek, Kyrgyzstan
    \and
    Gulira Jumalieva \\
    \textit{Dep. of Computer Linguistics} \\
    \textit{KSTU n. a. I. Razzakov}\\
    Bishkek, Kyrgyzstan
    \and
    Gulnara Kabaeva \\
    \textit{Information Technology Institute} \\
    \textit{KSTU n.a. I. Razzakov}\\
    Bishkek, Kyrgyzstan
    \and
    Sergey Nikolenko\\
    \textit{St. Petersburg Dep. of Steklov Math. Institute}\\\textit{St. Petersburg State University} \\
    St. Petersburg, Russia
}
\date{}

\maketitle
\begin{multicols}{2}

\begin{abstract}
% \scriptsize
% math symbols not allowed here! as well as special characters
We introduce KyrgyzNER, the first manually annotated named entity recognition dataset for the Kyrgyz language. Comprising 1{,}499 news articles from the 24.KG news portal, the dataset contains 10{,}900 sentences and 39{,}075 entity mentions across 27 named entity classes. We show our annotation scheme, discuss the challenges encountered in the annotation process, and present the descriptive statistics. We also evaluate several named entity recognition models, including traditional sequence labeling approaches based on conditional random fields and state-of-the-art multilingual transformer-based models fine-tuned on our dataset. While all models show difficulties with rare entity categories, models such as the multilingual RoBERTa variant pretrained on a large corpus across many languages achieve a promising balance between precision and recall. These findings emphasize both the challenges and opportunities of using multilingual pretrained models for processing languages with limited resources. Although the multilingual RoBERTa model performed best, other multilingual models yielded comparable results. This suggests that future work exploring more granular annotation schemes may offer deeper insights for Kyrgyz language processing pipelines evaluation.
\end{abstract}

\section{Introduction}\label{sec:intro}

Recent advances in machine learning and natural language processing (NLP) have been fueled by the availability of large, high-quality annotated datasets. However, the bulk of these resources have been developed for high-resource languages that have abundant linguistic data. \emph{Less-resourced languages} (LRL, low-resource languages) are those spoken in the world but with fewer linguistic resources for language technologies~\cite{cieri2016selection}; they are %significantly 
underrepresented in NLP research, and this imbalance not only limits the development of robust language technologies for LRL communities but also perpetuates inequalities in the access to cutting-edge NLP tools~\cite{hedderich2021survey}.
%
% The emergence of multilingual language models 
Multilingual LMs such as BERT~\cite{Devlin2019BERT} and XLM-RoBERTa (XLMR)~\cite{mt5}, trained on languages with varying amounts of resources, offer new possibilities for NLP in low-resource languages, e.g.,
% and provide
% . These models allow knowledge transfer from well-resourced languages to underrepresented ones, providing 
% a solid foundation 
for tasks such as \emph{named entity recognition} (NER), a core NLP task that involves identifying and classifying specific entities in text, such as names of people, locations, or organizations~\cite{jurafsky2008speech}, or other predefined domain-specific categories~\cite{miftahutdinov2020biomedical}. 

Kyrgyz is a prime example of a less-resourced language, with only a limited number of tools and datasets available. 
Developing manually annotated datasets for Kyrgyz is essential for evaluating and improving language models. While multilingual models continue to evolve, creating human-validated datasets remains a fundamental step in building reliable language resources. Even in the modern era of LLMs, 
it is hard to envision progress without %, at the very least, 
access to the evaluation data {prepared and/or validated by humans}. However, there are no high-quality manually annotated datasets for NER in Kyrgyz, which limits progress in this area.

% \begin{figure*}[!t]
%     \centering
%     \includegraphics[width=0.7\linewidth]{images/Grobid_ner.png}
%     % \caption{Caption}
%     % \label{fig:placeholder}
% \end{figure*}

In this work, we address this gap by presenting the first manually annotated Kyrgyz NER dataset, based on news articles collected from \url{https://24.kg/} with the permission of the 24.KG agency's editors to use the data for research purposes. Our contributions are as follows:
\begin{inparaenum}[(1)]
    \item we introduce a new dataset with 10{,}900 sentences and 39{,}075 entity mentions classified into 27 categories;
    \item we evaluate multiple baseline models, including classical and mainstream modern NER approaches;
    \item we establish a benchmark for Kyrgyz NER, providing a foundation for future research.
\end{inparaenum}
% The rest of the paper is organized as follows: 
Section~\ref{sec:relatedwork} reviews related work, Section~\ref{sec:corpus} presents the dataset and its annotation process, Section~\ref{sec:data_stats} reports corpus-related statistics, Section~\ref{sec:experiments} describes the baseline models, our experimental setup, and experimental results, Section~\ref{sec:error_discussion} investigates recurring mistake patterns, and Section~\ref{sec:conclusion} concludes the paper\footnote{The camera-ready version of this work has been accepted to the TurkLang-2025 conference, the paper's DOI and IEEE copyright will be added upon the confirmation of the acceptance for publication in the IEEE Xplore proceedings.}.

\section{Related Work}\label{sec:relatedwork}

\subsection{Named Entity Recognition}

BiLSTM-CRF models~\cite{vajjala2022reallyknowstateart} first dominated NER, later eclipsed by Transformer models like BERT~\cite{Devlin2019BERT}, whose contextual embeddings markedly improved multilingual performance. GPT-based systems now frame NER as text generation, achieving strong few-shot results but introducing hallucinations that call for self-verification~\cite{wang2023gptnernamedentityrecognition}.

\emph{Nested NER} identifies overlapping or hierarchical entities---e.g., ``The Chinese embassy in France,'' comprising facility and location spans~\cite{finkel2009nested}. Though valuable across domains~\cite{wang2020pyramid,loukachevitch2023nerel}, its complex annotation is beyond this study, which targets standard (``flat'') Kyrgyz NER.

\subsection{Datasets for Less-Resourced Languages}

Custom datasets are crucial for addressing the unique linguistic features of less-resourced languages, just like, e.g.,  
% and similar techniques have been used in other domains for similar problems. For instance, 
the biomedical domain has specialized corpora for disease and drug entity recognition for unique terminologies and nomenclatures of the field
% have driven advances in biomedical text mining
~\cite{Zhang2021MedicalNER}. This approach of building domain-specific datasets mirrors the development of a custom Kyrgyz NER dataset, which addresses the lack of language resources by capturing the specific linguistic characteristics of Kyrgyz.
Several widely used datasets have significantly advanced NER research and applications, including
\emph{CoNLL-2003}~\cite{tjong-kim-sang-de-meulder-2003-introduction},
\emph{OntoNotes 5.0}~\cite{pradhan-etal-2013-towards}, and
\emph{WNUT 2016/2017}~\cite{strauss-etal-2016-results}.
% \item 
\emph{WikiANN} is a large multilingual dataset annotated for named entities in over 282 languages, including support for less-resourced languages such as Uzbek, Turkish, Tatar, and Kazakh; it is a critical resource for low-resource NLP tasks~\cite{rahimi2019massivelymultilingualtransferner}.

\subsection{NER Datasets for Turkic Languages}

\begin{table*}[!t]
    \centering
    %\small\setlength{\tabcolsep}{2pt}
    \begin{tabular}{p{.16\linewidth}p{.1\linewidth}p{.11\linewidth}p{.07\linewidth}p{.13\linewidth}}
        \toprule
        % \hline
        \textbf{Dataset}      & \textbf{Language} & \textbf{\# Sentences} & \textbf{Classes} & $\quad$\textbf{Sources} \\
        \midrule
        % \hline
        KyrgyzNER & Kyrgyz & 10,900 & 27 & News (24.KG) \\ \hline
        Uzbek NER & Uzbek & 1,160 & 6 & News and social \\ \hline
        KazNERD & Kazakh & 112,702 & 25 & Wiki and news \\ \hline
        Turkish Wiki NER & Turkish & 20,000 & 3 & Wiki \\ \hline
        WikiANN & Multiple & Variable & 3 & Wiki (282 lang.) \\ 
        % \hline
        \bottomrule
    \end{tabular}
    \caption{Comparison of Turkic and Related NER Datasets}
    \label{tab:ner-comparison}
\end{table*}

Several datasets have been developed specifically for Turkic languages, including
\begin{inparaenum}[(i)]    
    \item \emph{Uzbek NER Dataset} with $1{,}160$ sentences annotated for parts of speech and named entities in the Uzbek language~\cite{Mengliev2024-ey};
    \item \emph{Kazakh NER Dataset} (KazNERD), an open-source dataset with $112{,}702$ sentences and $136{,}333$ annotations across 25 entity classes~\cite{yeshpanov-etal-2022-kaznerd};
    % ; it is available on GitHub and offers robust support for NER in Kazakh~\cite{yeshpanov-etal-2022-kaznerd};
    \item \emph{Turkish Wiki NER Dataset} with 20{,}000 sentences sampled and re-annotated from Kuzgunlar NER; this resource focuses on entity types such as \emph{Person}, \emph{Location}, and \emph{Organization}~\cite{altinok-2023-diverse}.
    % ; it is hosted on GitHub and Hugging Face.
\end{inparaenum}
For descriptive statistics, please see Table~\ref{tab:ner-comparison}.
The only NER model currently available in the Kyrgyz language is M.~Jumashev's \href{https://huggingface.co/murat/kyrgyz_language_NER}{kyrgyz\_language\_NER} model trained on \emph{WikiANN}, which is, according to the author himself, ``not usable''.

\subsection{SoTA in Named Entity Recognition}

NER research continues to evolve rapidly, driven by advancements in deep learning, with larger multilingual corpora and pre-trained models being made available and Transformer-based architectures continuously being improved~\cite{Lample2024Transformers}.
% such as RoBERTa and XLMR provide state-of-the-art performance by capturing semantic nuances and linguistic patterns across different languages. Recent models incorporate more sophisticated attention mechanisms and larger training corpora, leading to state-of-the-art results on various NER benchmarks~\cite{Lample2024Transformers}.
%
Recent innovations include integrating external knowledge bases into neural models, 
enabling better recognition of rare or unseen entities, e.g., 
% this hybrid approach has shown 
% a promising approach in domains such as 
in biomedical NER~\cite{Liu2024KnowledgeEnhancedNER}.
% , where new domain-specific entities frequently appear~\cite{Liu2024KnowledgeEnhancedNER}.
and zero-shot and few-shot learning techniques that 
% have also gained popularity, allowing models to 
let models generalize to new domains or languages with minimal training data and without extensive retraining at all~\cite{Brown2024ZeroShotNER}.
Cross-lingual and multilingual NER models have demonstrated significant potential in addressing low-resource challenges. NER models are increasingly being adapted to less-resourced languages
% and domains with specific terminologies such as medical or legal texts. Researchers are increasingly using
with transfer learning, domain adaptation, and few-shot learning.
% to enable models to generalize better from high-resource to low-resource settings. This has resulted in substantial improvements in the performance of 
In particular, one can improve NER systems for languages such as Kyrgyz by leveraging knowledge from more widely spoken languages such as English or Russian~\cite{Peters2024LowResourceNER}.
Finally, multilingual and cross-lingual NER models are also increasingly able to exploit shared linguistic features across languages~\cite{Conneau2024MultilingualNER}.
% to improve recognition accuracy for underrepresented languages such as Kyrgyz. By using shared subword tokenization and multilingual embeddings, these models have improved the accuracy and robustness of NER in languages with limited resources and varied orthographic systems~\cite{Conneau2024MultilingualNER} 
Overall, state of the art NER approaches in 2025 focus on advanced neural architectures, integrating external knowledge, and enhancing model adaptability to diverse languages and domains. This has resulted in more accurate, efficient, and scalable NER systems capable of operating effectively in multilingual and low-resource environments.

\subsection{Kyrgyz Language Processing}

Research on the Kyrgyz language has primarily focused on linguistic studies rather than computational approaches. A substantial body of linguistic research has already been devoted to various aspects of the Kyrgyz language; see a recent survey of Kyrgyz NLP research~\cite{alekseev2024kyrgyznlp}.
While projects promoting the Kyrgyz language, both commercial~\cite{akylai_24kg} and non-commercial~\cite{sexed_kyrgyz}, have increased recently, there remains a severe lack of annotated datasets for NLP tasks. With this work, we aim to address this gap by providing the first manually annotated dataset for Kyrgyz NER and offering baseline results to support future research.

\section{Kyrgyz NER Corpus}\label{sec:corpus}

In this section, we introduce the first manually annotated dataset for named entity recognition (NER) in Kyrgyz. Given the lack of pre-existing high-performance NER models for Kyrgyz, we explored two approaches to address this gap. The first approach involved building a NER model from scratch, while the second focused on adapting existing multilingual models by fine-tuning them on Kyrgyz data.

\subsection{Dataset}

The dataset consists of 1{,}499 news articles in Kyrgyz, collected from the 24.KG news portal with permission from the agency's editors for research purposes. These articles, dating from May 2017 to October 2022, were manually annotated to identify named entities using an annotation scheme adapted from the GROBID NER project~\cite{GROBID}. The dataset contains 10{,}900 sentences and 39{,}075 entity mentions across 27 classes, making it the most comprehensive resource for Kyrgyz NER. For annotation, we used the open-source tool \emph{Doccano}~\cite{doccano}.

\subsection{Annotation Support Tool}

The annotation process for named entities is labor-intensive and demands both linguistic expertise and consistent attention to detail. Annotation tools aim to improve the efficiency of this process while minimizing human error. After evaluating several options, we selected \emph{Doccano} as the primary tool for annotation due to its flexible interface and support for sequence labeling tasks. The annotation guidelines were adapted from GROBID (GeneRation Of BIbliographic Data) and customized to fit the specific requirements of the Kyrgyz language.

\subsection{Annotation Guideline}

We developed detailed annotation guidelines to ensure consistent and accurate labeling of entity mentions in the Kyrgyz NER dataset based on the guidelines from the GROBID project~\cite{GROBID}. Our tagset covers both broad and specific categories, capturing the diversity of named entities in Kyrgyz texts. Annotators were provided with practical examples and case studies to help them resolve common ambiguities. One of the most challenging cases involves context-dependent named entities. For instance, the word ``\ru{Президент}'' (President) can be labeled as either a \emph{Title} or a \emph{Person}, depending on the context:

\begin{enumerate}[(1)]
    \item \ru{$\langle$Президент Сооронбай Жээнбеков$\rangle$ бүгүн премьер-министр Сапар Исаковду кабыл алды.} \\
    \textit{$\langle$Prezident Sooronbay Jeenbekov$\rangle$ bügün premer-ministr Sapar İsakovdu kabıl aldı.} \\
    $\langle$President Sooronbai Jeenbekov$\rangle$ received Prime Minister Sapar Isakov today.\\
    In this case, the words \ru{``Президент Сооронбай Жээнбеков''} is a \emph{Person} named entity.
    \item \ru{$\langle$Президент$\rangle$ бүгүн премьер-министрди кабыл алды.} \\
    \textit{$\langle$Prezident$\rangle$ bügün premer-ministrdi kabıl aldı.} \\
    $\langle$President$\rangle$ received the Prime Minister today.\\
    In this case, \ru{``Президент''} is a \emph{Title}.
\end{enumerate}

Our guidelines follow the ``largest entity mention'' principle inherited from the GROBID project: in cases of nested entities, only the encompassing entity is annotated. For example, the token ``\ru{Кыргыз}'' (Kyrgyz) can be classified as \emph{National} (when referring to nationality), \emph{Person Type} (when referring to the Kyrgyz people), or \emph{Concept} (when referring to the Kyrgyz language), depending on the context.

{For more details and examples,} we refer to the ``largest entity mention'' section {in the original guidelines} available at \url{https://grobid-ner.readthedocs.io/en/latest/largest-entity-mention/} (as of mid-2025). The complete instructions of our own annotation scheme are provided in the Appendix.

\subsection{Annotators}

We trained $59$ native Kyrgyz speakers with relevant linguistic expertise as annotators. Working from detailed guidelines, they met regularly to resolve tricky cases and coordinated through an online channel with the annotation manager (one of the authors). Each document received multiple independent annotations; domain experts then compared versions, selected the most accurate spans, and served as final approvers to ensure consistency and accuracy.

\subsection{Annotation Process}

The annotation workflow was designed following the MATTER (Model, Annotate, Train, Test, Evaluate, and Revise) schema~\cite{pustejovsky2013natural} and other related work~\cite{foppiano2021supermat,otto2023gsapnernoveltaskcorpus,naraki2024augmentingnerdatasetsllms}. The workflow consisted of five steps, as illustrated in Fig.~\ref{fig:workflow}:
\begin{enumerate}[(1)]
\item \emph{data preparation}: the data (news articles) was prepared and uploaded to the annotation system, in our case a \emph{Doccano} instance~\cite{doccano};
    \item \emph{annotation}: a human annotator can select a document and manually add, remove, or modify each entity based on the instructions from the guidelines; once a document was fully annotated, it was marked as ``ready for validation'';
    \item \emph{validation/curation}: annotations from different users for a given document are validated and merged into a final annotation; a domain expert (``annotation approver'') can compare different annotated versions and select the best combination of annotations or add new ones; this step ensures that the annotations are cross-checked and that the document is validated by domain experts;
    \item \emph{consistency checks and statistical analysis}: this step focuses on identifying obvious errors such as mislabeled data or incorrect linkages; a sequence labeling model is trained and evaluated using 10-fold cross-validation, providing precision, recall, and F-score metrics for each label; on the next iteration, the model is used to automatically generate annotated data, and its predictions serve as a foundation for further refinement and analysis;
    \item \emph{review}: retrospective analysis of the iteration, where unclear cases are discussed and documented in the annotation guidelines.
\end{enumerate}

To inspect and further improve data quality, after the second update we trained a \emph{DeepPavlov NER}~\cite{savkin2024deeppavlov} model on the annotated data to see whether our data can be used to train the model and identify annotation errors. We split the dataset into 1{,}000 training texts and 500 test texts, achieving an F1 score of 66.16\% on the test set. This feedback loop allowed us to filter out additional errors and improve the dataset through multiple correction sessions.
As a result,
the inter-annotator agreement (Cohen's kappa) for $30$ sampled texts composed of $2{,}773$ tokens is
% . The agreement score is 
$\kappa = 0.89$, which is quite high and shows that we have a high-quality dataset.
% This rather high agreement score serves as evidence that we have obtained a high-quality dataset.

%%%%%%%%%%%%%%%%%%%%%%%%%%%%%%%%%%%%%%%%%%%%%%%%%%%%%%%%%

\begin{table*}[!t]
  \centering  
  \begin{minipage}[t]{0.43\textwidth}
    \vspace{-92pt}
    \begin{figure}[H]
      \centering
      \includegraphics[width=0.999\linewidth]{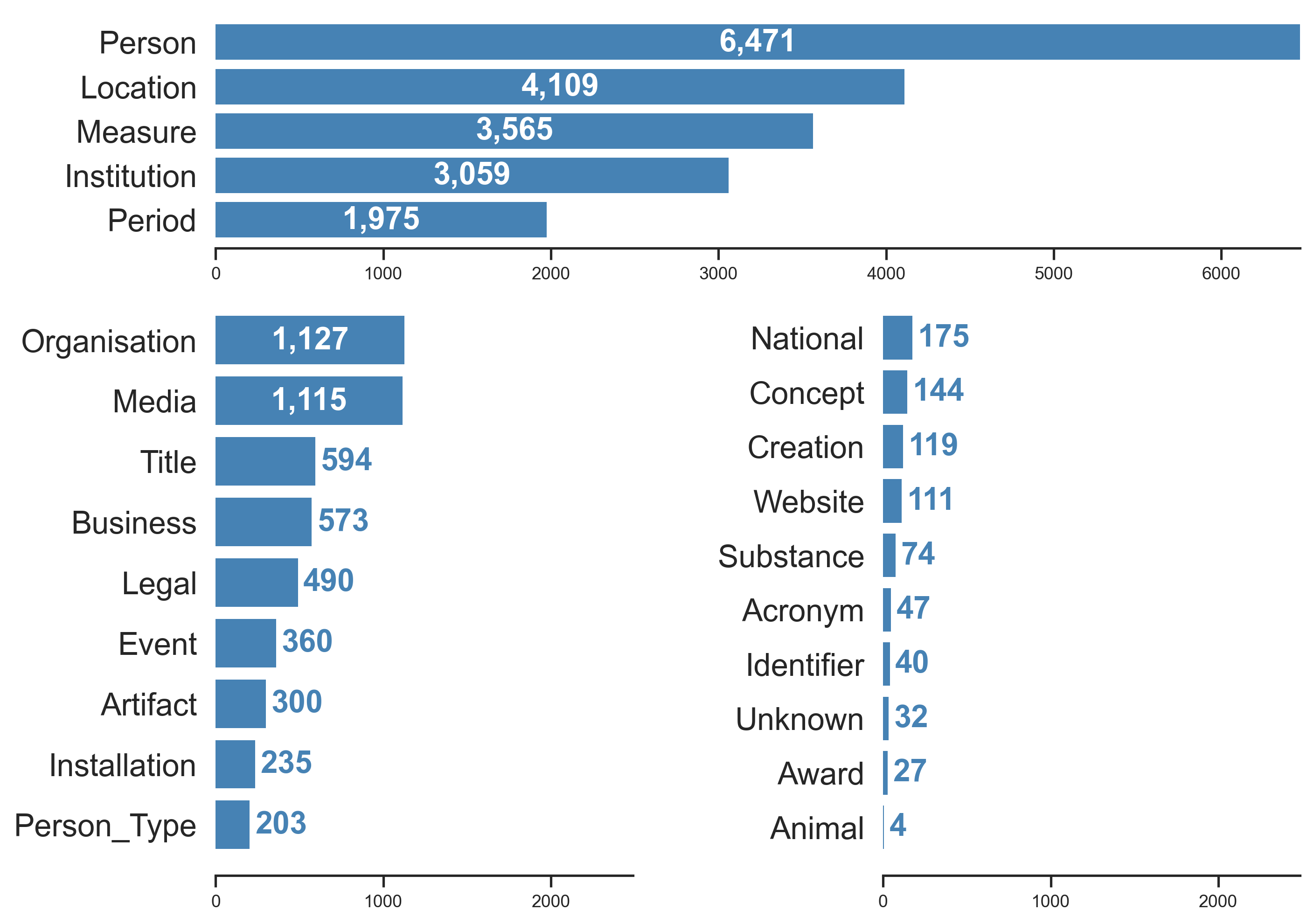}
      \caption{Distribution of label categories}
      \label{fig:distribution_classes}
    \end{figure}
  \end{minipage}%
  ~~~~~~~~%
  \begin{minipage}[t]{0.35\textwidth}
      \begin{tabular}{lccl}
        \toprule
         % \hline
        \ru{Апрель} & - & - & B-PERIOD \\
        \ru{айынана} & - & - & O \\
        \ru{баштап} & - & - & O \\
        \ru{мамлекеттик} & - & - & O \\
        \ru{жана} & - & - & O \\
        \ru{муниципалдык} & - & - & O \\
        \ru{кызматкерлер} & - & - & O \\
        \ru{кыргыз} & - & - & B-CONCEPT \\
        \ru{тилинен} & - & - & I-CONCEPT \\
        \ru{сынак} & - & - & O \\
        \ru{тапшырат} & - & - & O \\
        \ru{.} & - & - & O \\
        \bottomrule
         % \hline
      \end{tabular}
      \caption{Dataset in the CoNLL Format}
      \label{tab:conll_format}
  \end{minipage}
\end{table*}

% \begin{table*}[!t]\centering
%   \begin{tabular}{lccl}
%     % \toprule
%      \hline
%     \ru{Апрель} & - & - & B-PERIOD \\
%     \ru{айынана} & - & - & O \\
%     \ru{баштап} & - & - & O \\
%     \ru{мамлекеттик} & - & - & O \\
%     \ru{жана} & - & - & O \\
%     \ru{муниципалдык} & - & - & O \\
%     \ru{кызматкерлер} & - & - & O \\
%     \ru{кыргыз} & - & - & B-CONCEPT \\
%     \ru{тилинен} & - & - & I-CONCEPT \\
%     \ru{сынак} & - & - & O \\
%     \ru{тапшырат} & - & - & O \\
%     \ru{.} & - & - & O \\
%     % \bottomrule
%      \hline
%   \end{tabular}
%   \caption{Dataset in the CoNLL Format}
%   \label{tab:conll_format}
% \end{table*}

\subsection{Data Format}

Our dataset is presented in the CoNLL-2003 format, as shown in Table~\ref{tab:conll_format}. Word boundaries were found with a tokenizer from the \emph{Apertium-Kir} tool~\cite{washington2012finite} ensuring compatibility with standard NER evaluation frameworks and making our dataset more convenient for future research.

\begin{table*}[!t]
  \centering
  \begin{minipage}[t]{0.22\textwidth}
    \centering
    \begin{figure}[H]
      \centering
      \includegraphics[width=\linewidth]{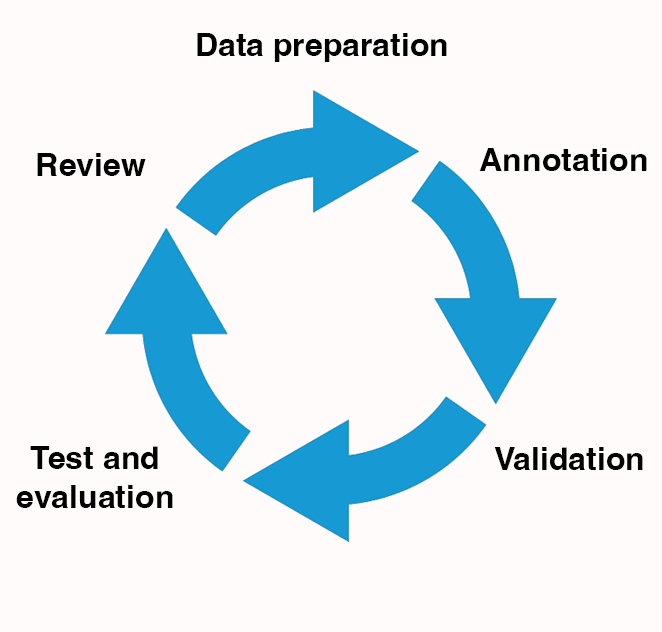}
      \caption{Workflow illustration.}\label{fig:workflow}
    \end{figure}
  \end{minipage}%
  % \hfill
  ~~~~%
\begin{minipage}[t]{0.37\textwidth}
    \centering
    \begin{table}[H]
      \centering
      \begin{tabular}{cccc}
        \toprule
         % \hline
        \textbf{Items} & \textbf{Train} & \textbf{Test} & \textbf{Total} \\ 
        % \hline
        \midrule
        \textbf{Documents} & $999$ & $500$ & $1{,}499$ \\ \hline
        \textbf{Sentences} & $7{,}033$ & $3{,}867$ & $10{,}900$ \\ \hline
        \textbf{Tokens} & $89{,}248$ & $51{,}118$ & $140{,}366$ \\ \hline
        \textbf{Mentions} & $24{,}949$ & $14{,}126$ & $39{,}075$ \\ 
        % \hline
        \bottomrule
      \end{tabular}
      \caption{Dataset Statistics.}
      \label{tab:data_stats}
    \end{table}
  \end{minipage}%
  % \hfill
~~~~%
\begin{minipage}[t]{0.33\textwidth}
    \centering
    \begin{table}[H]
      \centering
      \begin{tabular}{rccc}
        \toprule
         % \hline
        \textbf{Model} & \textbf{Prec} & \textbf{Rec} & \textbf{F1} \\ 
         % \hline
        \midrule
        \textbf{Bert+CRF} & 0.67 & 0.63 & 0.65 \\ \hline
        \textbf{CRF} & 0.70 & 0.55 & 0.62 \\  \hline
        \textbf{Pretrained mT5} & 0.70 & 0.68 & 0.69 \\  \hline
        \textbf{Pretrained Bert} & 0.68 & 0.68 & 0.68 \\  \hline
        \textbf{Pretrained XLMR} & 0.74 & 0.71 & \textbf{0.73} \\  
        % \hline
        \bottomrule
      \end{tabular}
      \caption{Results.}
      \label{tab:results}
    \end{table}
  \end{minipage}
\end{table*}

% \vspace{-8pt}main
\section{Data Statistics}\label{sec:data_stats}

In this section, we present an overview of the statistics and distribution of classes in the Kyrgyz NER dataset. The dataset comprises 1{,}499 documents, totaling 140{,}366 tokens and 10{,}900 sentences. The annotated dataset includes 39{,}075 named entity mentions distributed across 27 entity types. Table~\ref{tab:data_stats} provides dataset statistics and details its breakdown into training and test sets.
One of the key challenges for this dataset is the uneven distribution of entity classes. Fig.~\ref{fig:distribution_classes} shows that the four most frequent classes account for $\approx$70\% of all mentions, while many other classes have only a few samples. This class imbalance poses a significant challenge for training models, particularly for underrepresented classes. The histogram in Fig.~\ref{fig:distribution_classes} shows the frequency distribution of entity mentions for each class. Classes such as \emph{Person}, \emph{Location}, \emph{Measure}, and \emph{Institution} are among the most common, while others like \emph{Animal}, \emph{Award}, or \emph{Substance} are much rarer. The scarcity of examples for these underrepresented classes affects model performance and increases the likelihood of false negatives, and addressing class imbalance may be an important direction for future work.

\section{Experimental Setup and Results}\label{sec:experiments}

% In this section, we describe the experimental setup and results of our baseline models on the Kyrgyz NER dataset. 
The goals of the conducted experiments were:
\begin{inparaenum}[(1)]
\item to assess the performance of established NER techniques on the Kyrgyz dataset and identify the most suitable baseline for future research, and
\item to analyze how class imbalance affects model performance, given that the top four most frequent classes account for 70\% of mentions (see Fig.~\ref{fig:distribution_classes}).
\end{inparaenum}

\subsection{Baselines and Their Hyperparameters}

We split the dataset into training and validation subsets, using 20\% of the training set as a validation set.
We have trained a wide variety of models on the dataset, ranging from classical CRF-based models to state-of-the-art transformer-based models.

\textit{Classical baseline: CRF}. We utilized a Conditional Random Fields (CRF) model as the classical baseline for our experiments, training the model from scratch with the \emph{sklearn-crfsuite} library~\cite{CRFsuite} library. We used the L-BFGS optimization algorithm with a maximum of 5000 iterations to ensure convergence. To balance the trade-off between precision and generalization, we set the L1 regularization coefficient to 0.3 and the L2 regularization coefficient to 0.6. %Additionally, we enabled all possible state transitions to capture complex dependencies between labels.

\textit{BERT+CRF}. We used the \texttt{DeepPavlov} library~\cite{savkin2024deeppavlov} to train a model from scratch with the configuration ner\_ontonotes\_bert\_mult.%\footnote{\url{https://docs.deeppavlov.ai/en/master/features/models/NER.html#3.-Models-list}} 

\emph{BERT}. We fine-tuned BERT multilingual base (cased)~\cite{Devlin2019BERT} for NER in Kyrgyz. Fine-tuning was done with the \texttt{bert-base-multilingual-cased} checkpoint, with a batch size of $128$, learning rate of $5\cdot10^{-5}$, and weight decay $0.0001$. The model was trained for 8 epochs, with $3000$ warmup steps to stabilize the learning process. 

\emph{XLM-RoBERTa} (XLMR). We used the XMLR~\cite{conneau2020unsupervisedcrosslingualrepresentationlearning} base model to fine-tune the NER model for the Kyrgyz language. The fine-tuning process employed the xlm-roberta-base~\cite{conneau2020unsupervisedcrosslingualrepresentationlearning} checkpoint, with batch size $8$ and learning rate $10^{-5}$. To mitigate overfitting, we used a weight decay of $0.01$, training the model over $10$ epochs with $8000$ warmup steps.

\emph{mT5}. We fine-tuned the mT5-small model~\cite{mt5} to obtain a NER model for the Kyrgyz language. Fine-tuning used the google/mt5-small~\cite{mt5} checkpoint, with a batch size of 16 and a learning rate of $10^{-5}$. To prevent overfitting, we applied a weight decay of $0.001$. Training was conducted over $10$ epochs, with $800$ warmup steps to stabilize the learning process. The maximum token length was set to 64.

\begin{table*}[!ht]
  % \vspace{5pt}
  \centering
    % \setlength{\tabcolsep}{4.5pt}
  % \resizebox{\linewidth}{!}{
\begin{tabular}{rcccccc}
\toprule
\textbf{Label} & \textbf{BERT} & \textbf{CRF} & \textbf{mT5} & \textbf{BERT} & \textbf{XLMR} & \textbf{\#} \\
& \textbf{+CRF} & & & & &  \\  
\midrule
\emph{Measure} & 0.82 & 0.76 & 0.85 & 0.82 & 0.86 & 1046 \\  \hline
\emph{Person} & 0.80 & 0.66 & 0.82 & 0.77 & 0.82 & 989 \\  \hline
\emph{Location} & 0.74 & 0.66 & 0.76 & 0.73 & 0.76 & 900 \\  \hline
\emph{Institution} & 0.59 & 0.54 & 0.63 & 0.60 & 0.64 & 660 \\  \hline
\emph{Period} & 0.54 & 0.64 & 0.68 & 0.69 & 0.69 & 545 \\ %\midrule
 \hline
\emph{Plant} & 0.00 & 0.00 & 0.00 & 0.00 & 0.00 & 10 \\  \hline
\emph{Award} & 0.00 & 0.00 & 0.00 & 0.18 & 0.00 & 7 \\  \hline
\emph{Conceptual} & 0.00 & 0.00 & 0.00 & 0.00 & 0.00 & 5 \\  \hline
\emph{Identifier} & 0.00 & 0.00 & 0.00 & 0.42 & 0.67 & 4 \\  \hline
\emph{Animal} & 0.00 & 0.00 & 0.00 & 0.00 & 0.00 & 3 \\ 
\bottomrule
\end{tabular}

    \caption{Per-Class F1 Score}\label{tab:class}
% }
\end{table*}

\subsection{Experimental Results}

The results of our experiments are summarized in Table~\ref{tab:results}. As expected, the CRF model achieved high precision but struggled with recall, leading to a relatively low F1 score even after extensive hyperparameter tuning. Transformer-based models consistently outperformed the CRF baseline, with XLMR delivering the best overall performance (F1 score of 0.70). The mT5 model also performed well, indicating the potential of text-to-text Transformer-based models for low-resource NER tasks. We note that in a recent study on Kyrgyz texts classification~\cite{alekseev2023benchmarking}, the ``Large'' modification of XLMR also yielded the best results, while multilingual BERT was far behind. In this case, the performance gap was relatively small, so all Transformer-based models represent strong baselines for future studies on Kyrgyz tagging tasks.

Table~\ref{tab:class} shows a breakdown of the F1 score for several popular and rare entity classes; as expected, classes with very low support are virtually unrecognizable for the models, while larger amount of training data leads to significantly improved results for all models. Note the difference in results between \emph{Measure}/\emph{Person} and \emph{Institution}/\emph{Period} classes: as support drops from 900-1000 examples to 550-650, the F1 scores go down from 0.8-0.85 to 0.6-0.65.

%%%%%%%%%%%%%%%%%%%%%%%%%%%%%%%%%%%%%%%%%%%%%%%%%%%%%%%%%%%%%%%%%%%%%%%%

\section{Error Analysis}\label{sec:error_discussion}

We analyzed the predictions made by the BERT+CRF model to identify recurring mistake patterns. The findings highlight several critical issues, e.~g. ambiguous entity mentions and the scarcity of training examples for certain classes. As a result, we refined our annotation guidelines
% , and it was part of the annotation process 
(see Section~\ref{sec:corpus}). As for the baseline models trained on the resulting KyrgyzNER dataset, we identify two major sources of errors.

\emph{Ambiguous entity mentions}: one of the main sources of errors was context-dependent entity mentions that could belong to multiple classes. For example, the word ``British'' can be labeled as \emph{National} (``a British newspaper reported''), \emph{Person Type} (``a British journalist reported''), or \emph{Concept} (``a journalist reported in British English''), depending on the context. The model struggled to disambiguate these cases, frequently misclassifying them or producing false negatives. 

\emph{Scarcity of training samples}: another major challenge was the lack of sufficient training samples for certain classes, including \emph{Acronym}, \emph{Animal}, \emph{Artifact}, \emph{Award}, \emph{Concept}, \emph{Event}, \emph{Identifier}, \emph{Installation}, \emph{Legal}, \emph{Plant}, and \emph{Substance}. As a result, models often failed to predict any labels for these classes, leading to a high rate of false negatives. To address these challenges, we propose to either revise and filter the set of classes or extend the training data with either synthetic sentences or upsampling.

%%%%%%%%%%%%%%%%%%%%%%%%%%%%%%%%%%%%%%%%%%%%%%%%%%%%%%%%%%%%%%%%%%%%%%%%

\section{Conclusion}\label{sec:conclusion}

In this work, we present KyrgyzNER, the first manually annotated named entity recognition (NER) dataset for the Kyrgyz language. The dataset comprises 1{,}499 documents with 10{,}900 sentences and 39{,}075 entity mentions across 27 categories. This resource addresses the lack of available language datasets for Kyrgyz and aims to provide a foundation for further research in low-resource language processing and exploring the effects of an unbalanced label distribution.

We conducted baseline experiments with NER models ranging from classical CRF to modern Transformer-based approaches,
% such as multilingual BERT, XLMR, and mT5. Our results showed 
showing that Transformer-based models significantly outperform classical methods, with XLMR achieving the highest F1 score. However, class imbalance remains a major challenge, especially for rare entity types, and
% Interestingly, the scores achieved with 
scores of mBERT (base), XLMR (base), and mT5 are very close (F1 about 70\%), so we suggest that all of them should be used as baselines for future research.
Our error analysis revealed two key issues: ambiguity in entity mentions and the scarcity of training samples for certain classes. To mitigate these challenges, we suggest refining annotation guidelines, generating synthetic data, and applying upsampling techniques in future work. These steps could improve model performance and provide a more robust benchmark for Kyrgyz NER.

We hope releasing this dataset spurs the NLP community to add Kyrgyz to multilingual benchmarks and deepen low-resource research, thereby expanding resources and narrowing the gap between high- and low-resource languages.

\section{Limitations}

KyrgyzNER faces four key constraints: 
\begin{inparaenum}[(1)]
\item its texts come exclusively from the 24.KG news portal, so models may fail to generalize to other genres such as social media, conversations, or legal prose; 
\item class imbalance remains severe: \emph{Person}, \emph{Location}, and \emph{Measure} are plentiful, whereas \emph{Acronym}, \emph{Animal}, \emph{Award}, and \emph{Legal} have few instances, raising false-negative risk; adding varied sources would help
\item the scheme annotates only ``flat'' entities, omitting nested or overlapping mentions, which limits certain applications;
\item although rigorous checks produced a strong Cohen's kappa of $0.89$, nuanced Kyrgyz usages still allow missed or inconsistent labels, so richer guidelines and further validation are needed.
\end{inparaenum}

\section*{Acknowledgement} % да, без s

We thank KSTU n.~a. I.~Razzakov students, teachers, and all other volunteers who participated in the annotation effort. Contributors are listed on page: \url{https://github.com/Akyl-AI/KyrgyzNER/blob/main/volunteers.md}. The authors also thank the anonymous reviewers whose comments have allowed us to improve the paper. A.~A. thanks Russian Science Foundation grant \# 23-11-00358 for support.

\bibliographystyle{plain}
% \bibliography{output}

\begin{thebibliography}{10}

\bibitem{GROBID}
Grobid.
\newblock \url{https://github.com/kermitt2/grobid}, 2008--2025.

\bibitem{alekseev2023benchmarking}
Anton Alekseev, Sergey Nikolenko, and Gulnara Kabaeva.
\newblock Benchmarking multilabel topic classification in the {K}yrgyz language.
\newblock In {\em International Conference on Analysis of Images, Social Networks and Texts}, pages 21--35. Springer, 2023.

\bibitem{alekseev2024kyrgyznlp}
Anton Alekseev and Timur Turatali.
\newblock Kyrgyznlp: Challenges, progress, and future.
\newblock In {\em International Conference on Analysis of Images, Social Networks and Texts}, pages 3--39. Springer, 2024.

\bibitem{altinok-2023-diverse}
Duygu Altinok.
\newblock A diverse set of freely available linguistic resources for {T}urkish.
\newblock In {\em Proceedings of the 61st Annual Meeting of the Association for Computational Linguistics}, pages 13739--13750, Toronto, Canada, July 2023. Association for Computational Linguistics.

\bibitem{Brown2024ZeroShotNER}
Tom Brown, Benjamin Mann, Nick Ryder, Melanie Subbiah, Jared~D Kaplan, et~al.
\newblock Language models are few-shot learners.
\newblock In {\em Proceedings of the 2024 Annual Conference on Neural Information Processing Systems (NeurIPS)}. Curran Associates, Inc., 2024.

\bibitem{cieri2016selection}
Christopher Cieri, Mike Maxwell, Stephanie Strassel, and Jennifer Tracey.
\newblock Selection criteria for low resource language programs.
\newblock In {\em Proceedings of LREC'16}, pages 4543--4549, 2016.

\bibitem{conneau2020unsupervisedcrosslingualrepresentationlearning}
Alexis Conneau, Kartikay Khandelwal, Naman Goyal, Vishrav Chaudhary, Guillaume Wenzek, et~al.
\newblock Unsupervised cross-lingual representation learning at scale.
\newblock In {\em Proceedings of the 58th Annual Meeting of the Association for Computational Linguistics}, page 8440. Association for Computational Linguistics, 2020.

\bibitem{Conneau2024MultilingualNER}
Alexis Conneau and Guillaume Lample.
\newblock Cross-lingual language model pretraining.
\newblock {\em Proceedings of the 2024 Conference on Empirical Methods in Natural Language Processing (EMNLP)}, 2024.

\bibitem{Devlin2019BERT}
Jacob Devlin, Ming-Wei Chang, Kenton Lee, and Kristina Toutanova.
\newblock {BERT}: Pre-training of deep bidirectional transformers for language understanding.
\newblock {\em Proceedings of NAACL-HLT-2019}, pages 4171--4186, 2019.

\bibitem{finkel2009nested}
Jenny~Rose Finkel and Christopher~D Manning.
\newblock Nested named entity recognition.
\newblock In {\em Proceedings of EMNLP 2009 conference}, pages 141--150, 2009.

\bibitem{foppiano2021supermat}
Luca Foppiano, Sae Dieb, Akira Suzuki, Pedro Baptista~de Castro, Suguru Iwasaki, et~al.
\newblock Supermat: construction of a linked annotated dataset from superconductors-related publications.
\newblock {\em Science and Technology of Advanced Materials: Methods}, 1(1):34--44, 2021.

\bibitem{hedderich2021survey}
Michael~A Hedderich, Lukas Lange, Heike Adel, Jannik Str{\"o}tgen, and Dietrich Klakow.
\newblock A survey on recent approaches for natural language processing in low-resource scenarios.
\newblock In {\em Proceedings of the 2021 Conference of the North American Chapter of the Association for Computational Linguistics: Human Language Technologies}, pages 2545--2568, 2021.

\bibitem{jurafsky2008speech}
Daniel Jurafsky and James~H Martin.
\newblock Speech and language processing. prentice hall.
\newblock {\em Upper Saddle River, NJ}, 2008.

\bibitem{akylai_24kg}
Anastasia Kan.
\newblock {AkylAI} smart speaker: Artificial intelligence speaking {Kyrgyz} language (june 18th, 2024).
\newblock \url{https://web.archive.org/web/20240619010036/https://24.kg/english/296874_AkylAI_smart_speaker_Artificial_intelligence_speaking_Kyrgyz_language/}, 2024.
\newblock Accessed: 2024-09-14.

\bibitem{Lample2024Transformers}
Guillaume Lample, Alexis Conneau, Ludovic Denoyer, and Marc'Aurelio Ranzato.
\newblock Unsupervised machine translation using monolingual corpora only.
\newblock In {\em Proceedings of the 2024 Annual Conference on Neural Information Processing Systems (NeurIPS)}. Curran Associates, Inc., 2024.

\bibitem{Liu2024KnowledgeEnhancedNER}
Tianyu Liu, Jin-Ge Yao, and Chin-Yew Lin.
\newblock Towards improving neural named entity recognition with gazetteers.
\newblock {\em Journal of Machine Learning Research}, 25:1--29, 2024.

\bibitem{loukachevitch2023nerel}
Natalia Loukachevitch, Suresh Manandhar, Elina Baral, Igor Rozhkov, Pavel Braslavski, et~al.
\newblock Nerel-bio: a dataset of biomedical abstracts annotated with nested named entities.
\newblock {\em Bioinformatics}, 39(4):btad161, 2023.

\bibitem{Mengliev2024-ey}
Davlatyor Mengliev, Vladimir Barakhnin, Nilufar Abdurakhmonova, and Mukhriddin Eshkulov.
\newblock Developing named entity recognition algorithms for uzbek: Dataset insights and implementation.
\newblock {\em Data Brief}, 54:110413, June 2024.

\bibitem{miftahutdinov2020biomedical}
Zulfat Miftahutdinov, Ilseyar Alimova, and Elena Tutubalina.
\newblock On biomedical named entity recognition: experiments in interlingual transfer for clinical and social media texts.
\newblock In {\em ECIR}, pages 281--288. Springer, 2020.

\bibitem{doccano}
Hiroki Nakayama, Takahiro Kubo, Junya Kamura, Yasufumi Taniguchi, and Xu~Liang.
\newblock doccano: Text annotation tool for human, 2018.
\newblock Available from https://github.com/doccano/doccano.

\bibitem{naraki2024augmentingnerdatasetsllms}
Yuji Naraki, Ryosuke Yamaki, Yoshikazu Ikeda, Takafumi Horie, Kotaro Yoshida, Ryotaro Shimizu, et~al.
\newblock Augmenting ner datasets with llms: Towards automated and refined annotation, 2024.

\bibitem{CRFsuite}
Naoaki Okazaki.
\newblock Crfsuite: a fast implementation of conditional random fields (crfs), 2007.

\bibitem{otto2023gsapnernoveltaskcorpus}
Wolfgang Otto, Matth{\"a}us Zloch, Lu~Gan, Saurav Karmakar, and Stefan Dietze.
\newblock Gsap-ner: A novel task, corpus, and baseline for scholarly entity extraction focused on machine learning models and datasets, 2023.

\bibitem{Peters2024LowResourceNER}
Matthew~E Peters, Waleed Ammar, Chandra Bhagavatula, and Russell Power.
\newblock Semi-supervised sequence tagging with bidirectional language models.
\newblock {\em Transactions of the Association for Computational Linguistics}, 12:193--208, 2024.

\bibitem{pradhan-etal-2013-towards}
Sameer Pradhan, Alessandro Moschitti, Nianwen Xue, Hwee~Tou Ng, Anders Bj{\"o}rkelund, et~al.
\newblock Towards robust linguistic analysis using {O}nto{N}otes.
\newblock In {\em Proceedings of the Seventeenth Conference on Computational Natural Language Learning}, pages 143--152, Sofia, Bulgaria, August 2013. Association for Computational Linguistics.

\bibitem{pustejovsky2013natural}
J.~Pustejovsky and A.~Stubbs.
\newblock {\em Natural Language Annotation for Machine Learning}.
\newblock Number v. 9, in A Guide to corpus-building for applications. O'Reilly Media, Inc., 2013.

\bibitem{rahimi2019massivelymultilingualtransferner}
Afshin Rahimi, Yuan Li, and Trevor Cohn.
\newblock Massively multilingual transfer for ner.
\newblock In {\em Proceedings of the 57th Annual Meeting of the Association for Computational Linguistics}, pages 151--164, 2019.

\bibitem{savkin2024deeppavlov}
Maksim Savkin, Anastasia Voznyuk, Fedor Ignatov, Anna Korzanova, Dmitry Karpov, et~al.
\newblock Deeppavlov 1.0: Your gateway to advanced nlp models backed by transformers and transfer learning.
\newblock In {\em Proceedings of the 2024 Conference on Empirical Methods in Natural Language Processing: System Demonstrations}, pages 465--474, 2024.

\bibitem{strauss-etal-2016-results}
Benjamin Strauss, Bethany Toma, Alan Ritter, Marie-Catherine De~Marneffe, and Wei Xu.
\newblock Results of the {WNUT}16 named entity recognition shared task.
\newblock In {\em Proceedings of the 2nd Workshop on Noisy User-generated Text ({WNUT})}, pages 138--144, Osaka, Japan, December 2016. The COLING 2016 Organizing Committee.

\bibitem{tjong-kim-sang-de-meulder-2003-introduction}
Erik~F. Tjong Kim~Sang and Fien De~Meulder.
\newblock Introduction to the {C}o{NLL}-2003 shared task: Language-independent named entity recognition.
\newblock In {\em Proceedings of the Seventh Conference on Natural Language Learning at {HLT}-{NAACL} 2003}, pages 142--147, 2003.

\bibitem{sexed_kyrgyz}
UNESCO-IITE.
\newblock A chatbot for teenagers about puberty, relationships, and health launched in kyrgyzstan (may 24th, 2022).
\newblock \url{http://web.archive.org/web/20240525072322/https://iite.unesco.org/highlights/oilo-chatbot-sex-ed-kyrgyzstan-en/}, 2022.
\newblock Accessed: 2024-09-14.

\bibitem{vajjala2022reallyknowstateart}
Sowmya Vajjala and Ramya Balasubramaniam.
\newblock What do we really know about state of the art ner?, 2022.

\bibitem{wang2020pyramid}
Jue Wang, Lidan Shou, Ke~Chen, and Gang Chen.
\newblock Pyramid: A layered model for nested named entity recognition.
\newblock In {\em Proceedings of the 58th annual meeting of the association for computational linguistics}, pages 5918--5928, 2020.

\bibitem{wang2023gptnernamedentityrecognition}
Shuhe Wang, Xiaofei Sun, Xiaoya Li, Rongbin Ouyang, Fei Wu, et~al.
\newblock Gpt-ner: Named entity recognition via large language models.
\newblock In {\em Findings of the Association for Computational Linguistics: NAACL 2025}, pages 4257--4275, 2025.

\bibitem{washington2012finite}
Jonathan~North Washington, Mirlan Ipasov, and Francis~M Tyers.
\newblock A finite-state morphological transducer for kyrgyz.
\newblock In {\em LREC}, pages 934--940, 2012.

\bibitem{mt5}
Linting Xue, Noah Constant, Adam Roberts, Mihir Kale, Rami Al-Rfou, et~al.
\newblock m{T}5: A massively multilingual pre-trained text-to-text transformer.
\newblock In {\em Proceedings of the NAACL-HLT 2021 Conference}, pages 483--498. Association for Computational Linguistics, 2021.

\bibitem{yeshpanov-etal-2022-kaznerd}
Rustem Yeshpanov, Yerbolat Khassanov, and Huseyin~Atakan Varol.
\newblock {K}az{NERD}: {K}azakh named entity recognition dataset.
\newblock In {\em Proceedings of the Thirteenth Language Resources and Evaluation Conference}, pages 417--426, Marseille, France, 2022. European Language Resources Association.

\bibitem{Zhang2021MedicalNER}
L.~Zhang and H.~Wu.
\newblock Medical text entity recognition based on deep learning.
\newblock {\em Journal of Physics: Conference Series}, 1744(4):042209, 2021.

\end{thebibliography}

\end{multicols}
\clearpage

\setlength{\parskip}{0.2\baselineskip}%
\setlength{\parindent}{0pt}%

\section*{Appendix: Guidelines for Entity Annotation}

This document outlines key principles and guidelines for entity annotation, providing definitions, examples, and addressing frequently asked questions. It covers various entity types such as persons, organizations, locations, and events, emphasizing consistency and precision in tagging\footnote{The original instructions have been prepared in Russian. Below is the translated version.}.

\section{Key Principles}
The ultimate goal of such methods is to extract some \textbf{useful} information from the text (not necessarily commercial, for example, for distant reading of the national epic by philologists); data should be annotated with this in mind.

\textbf{Example:} ``The policeman saved the boy.''~--- there's nothing to extract here, as we can't match any specific person to either the policeman or the boy in this text. However, if the unique title such as the ``Mayor of Bishkek'' is mentioned, we can do this.

The same goes for time intervals. ``Yesterday'' tells us nothing, but ``in January'' already allows us to make some intelligent guesses about the mentioned time span.

\subsection{The Principle of the Largest Entity}
The main rule: if there are nested entities, one should always take the largest one covering (including) everything (largest entity mention).

Consider the word ``British.'' Depending on the context, ``British'' may correspond to classes (labels):
\begin{itemize}
 \item \textbf{NATIONAL} (when referring to something related to the UK),
 \item \textbf{PERSON\_TYPE} (when referring to the British people),
 \item \textbf{CONCEPT} (when it means British English).
\end{itemize}

However, ``British Brexit referendum'' should be marked as an \textbf{EVENT} because ``British'' is a part of a larger entity.

\section{TITLE and PERSON}
If a title (position) is followed by a name, mark it as a single \textbf{PERSON} tag.

\textbf{Other examples:}
\begin{itemize}
 \item While attending the May 2012 NATO summit meeting (\textbf{EVENT}),
 \item Obama (\textbf{PERSON}) was the US President (\textbf{TITLE}),
 \item She is the CEO of IBM (\textbf{TITLE}),
 \item The President of Argentina (\textbf{TITLE}) said no,
 \item German South-West Africa (\textbf{LOCATION}),
 \item American Jewish Holocaust survivors (\textbf{PERSON\_TYPE}),
 \item chairman of the Central Committee of the World Sephardi Federation and a member of the Knesset (\textbf{TITLE}).
\end{itemize}

\textbf{IMPORTANT:} Entities should \textbf{NOT} be marked by clicking, to avoid space characters being included. The word should be selected \textbf{FULLY}; \textbf{PARTIAL} word selection is \textbf{NOT} allowed.

\section{Labels, Their Meaning, Examples}

\begin{table*}[!t]
\caption{Description of labels.}
\label{tab1}
\centering
\begin{tabular}{|p{3cm}|p{5cm}|p{3cm}|p{3cm}|}
\hline
\textbf{Label Name} & \textbf{Description} & \textbf{Examples EN} & \textbf{Examples KG} \\
\hline
PERSON       & Names, surnames, nicknames, and callsigns of real and fictional PEOPLE & John Smith & \ru{Исхак Раззаков, Чыңгыз Айтматов} \\ % \hline
PERSON\_TYPE & Type of person, societal role, often based on group membership       & African-American, Asian, Conservative, Liberal, Jews & \ru{кыргыз, татар, түрк} \\ %\hline
ANIMAL       & Animal names                                                          & Hachikō, Jappeloup & \ru{Хатико} \\ % \hline
TITLE        & Title, professional address, or position                             & Mr., Dr., General, President & \ru{Мырза, Президент} \\  %\hline
ORGANISATION & Organized group of people, with some form of legal entity and concrete membership & Alcoholics Anonymous, Jewish resistance, Polish underground & \ru{Кыргыз кино} \\  %\hline
INSTITUTION  & Organization of people and a location or structure that shares the same name & Yale University, European Patent Office, the British government & \ru{КГТУ, Политех, Кыргызпатент} \\  %\hline
BUSINESS     & A company or commercial organization                                 & Air Canada, Microsoft & \ru{Шоро} \\  %\hline
SPORT\_TEAM  & Organization or group associated with sports                          & The Yankees, Leicestershire & \ru{ФК Дордой-динамо} \\  %\hline
MEDIA        & Media, publishing organization, or name of the publication itself     & Le Monde, The New York Times & \ru{Кактус медиа, Клооп, 24.KG} \\  %\hline
WEBSITE      & Website name or link. For news companies, mark as MEDIA.             & Wikipedia, http://www.inria.fr & Instagram, Facebook \\  %\hline
IDENTIFIER   & Identifier like phone number, email, ISBN                            & 2081396505, weirdturtle@gmail.com & +996312000001 \\ % \hline
LOCATION     & A specific place; includes planets and galaxies                      & Los Angeles, Earth & \ru{Кыргызстан, Бишкек, Чолпон} \\ % \hline
NATIONAL     & Pertaining to location or nationality                                & a British newspaper, a British historian & \ru{кыргызстандык, орусиялык} \\  %\hline
INSTALLATION & A structure built by people                                          & Strasbourg Cathedral, Auschwitz camp & \ru{Бурана} \\  %\hline
ARTIFACT     & A man-made object, including software products                       & FIAT 634, Microsoft Word & \ru{«Гиннес китеби»} \\ % \hline
CREATION     & A work of art or entertainment, such as a song, movie, or book       & Mona Lisa, Kitchen Nightmares & \ru{Дочь Советской Киргизии} \\ % \hline
EVENT        & A specific event                                                     & World War 2, Brexit referendum & \ru{Экинчи дуйнолук согуш} \\  %\hline
AWARD        & Award for achievements in various fields                             & Ballon d'Or, Nobel Prize & \ru{Нобель сыйлыгы} \\  %\hline
PERIOD       & Date or historical period, time intervals                            & January, 1985-1989 & \ru{15 март, дүйшөмбү} \\  %\hline
LEGAL        & Legal references like laws, conventions, court cases                 & European Patent Convention, Roe v. Wade & \\  %\hline
MEASURE      & Numerical or ordinal value                                           & 1,500, 72\% & \\  %\hline
PLANT        & Name of a plant                                                      & Ficus religiosa & \\  %\hline
SUBSTANCE    & Substance                                                            & HCN, gold & \ru{алтын, күмүш} \\  %\hline
CONCEPT      & Abstract concept, not included in any other class                    & Communism, FTSE 100 & \ru{Коммунизм, кыргыз тили} \\  %\hline
CONCEPTUAL   & Entity associated with a concept                                     & Greek myths, eurosceptic doctrine & \\ % \hline
ACRONYM      & Acronyms and abbreviations                                           & DIY, BYOD, IMHO & \\  %\hline
UNKNOWN      & Entity that does not fit into any listed classes                     & Plan Marshall, Horizon 2020 & \ru{Маршалл планы} \\ % \hline
\hline
\end{tabular}
\end{table*}

\subsection{Frequently Asked Questions}

\begin{itemize}
    \item \textbf{Do quotes around an organization or sports team name get included?} Yes, the quotes should be marked along with the entity.
    \item \textbf{Can an entity be split into parts?} No, it cannot.
    \item \textbf{Is a road, an interchange, or a ring road considered INSTALLATION or LOCATION?} We'll consider it as LOCATION.
    \item \textbf{What label should be used for a specific hospital, prison, school, theater, or border crossing?} These should be marked as INSTITUTIONs.
    \item \textbf{What about a state-owned enterprise, like a specific factory?} This should be marked as BUSINESS.
    \item \textbf{If I encounter a difficult case and I am unsure what to do?} Gather a small batch of questions and post them in the speacial annotation-related chat [link].
\end{itemize}

\subsection{Clarifications}

\textbf{Specific Cases}
\begin{itemize}
    \item \textbf{``President Almazbek Atambayev''} is a single \textbf{PERSON} segment. If there is no name after ``President'', it should be labeled as a \textbf{TITLE}.
    \item \textbf{MEASURE} \\
    ``\ru{9 миллион 500 миң}'' (nine million five hundred thousand) should be marked as a single MEASURE segment (since it's a single number).
    \item Examples like ``\ru{10 сом}'' or ``5 apples''~--- only the number should be marked.
\end{itemize}

\subsection{Iteration \#1 (June 5-8, 2023)}

\textbf{GENERAL RULE}: If an entity becomes too long, please check whether it has been formed correctly; ensure that no sequences of entities are included into the spans, and no spaces or verbs are included into them.

If the same entity appears multiple times in the text, it should be labeled \textbf{each} time. Words should not be split into parts by selected spans: ``\ru{\textbf{70тен}}''

\subsection{Iteration \#2 (Questions and Answers After the First Month of Annotation Process)}

\begin{itemize}
    \item \textbf{Muftiate}~--- INSTITUTION.
    \item \textbf{Kyrgyz language}~--- should mark only  ```Kyrgyz''' as a CONCEPT.
    \item \textbf{Central Mosque in Bishkek}~--- INSTALLATION.
    \item \textbf{Religious Affairs Directorate}~--- INSTITUTION.
    \item \textbf{Religious Affairs Information and Counseling Center}~--- INSTITUTION.
    \item \textbf{Markets} (e.g., Osh Bazaar, Dordoi Bazaar)~--- BUSINESS.
    \item \textbf{Media abbreviation (\ru{ЖМК})}~--- ACRONYM.
    \item \textbf{Uzbekistan Gymnastics Federation}~--- ORGANISATION.
    \item \textbf{Rogun HPP} (Hydroelectric Power Plant)~--- INSTALLATION.
    \item \textbf{Makarov-type pistol}~--- ARTIFACT.
    \item \textbf{World Bank}~--- INSTITUTION.
    \item \textbf{``SDPK Party''}~--- ORGANISATION.
    \item \textbf{``Kyrgyzstan'' political party}~--- ORGANISATION.
    \item \textbf{Central Mosque in Bishkek}~--- INSTALLATION.
    \item \textbf{Manas (airport)}~--- INSTALLATION.
    \item \textbf{Tokmok city}~--- only mark ``TOKMOK'' part as a LOCATION.
\end{itemize}

\end{document}